\documentclass{article}

\usepackage{longcat_style}
\usepackage[export]{adjustbox}
\usepackage[utf8]{inputenc} %
\usepackage[T1]{fontenc}    %
\usepackage{newunicodechar}
\usepackage{hyperref}       %
\usepackage{xcolor}
\usepackage[normalem]{ulem} %
\hypersetup{
    colorlinks=true,      %
    linkcolor=blue,      %
    urlcolor=blue,       %
    citecolor=blue,      %
    linkbordercolor=blue, %
    urlbordercolor=blue,
    citebordercolor=blue,
    pdfborderstyle={/S/U/W 1}, %
}

\usepackage{float}
\usepackage{url}            %
\usepackage{booktabs}       %
\usepackage{amsfonts}       %
\usepackage{nicefrac}       %
\usepackage{microtype}      %
\usepackage{lipsum}		%
\usepackage{graphicx}
\usepackage{natbib}
\usepackage{doi}
\usepackage{amsmath}
\usepackage{amssymb} %
\usepackage{xspace}
\usepackage{enumitem}
\usepackage{multirow}
\usepackage{subcaption} 
\usepackage{makecell}
\usepackage{hyperref, cleveref}
\usepackage{pifont}
\usepackage[inkscapelatex=false]{svg}
\usepackage{caption}
\usepackage{wrapfig}
\usepackage{algorithm}
\usepackage{algorithmic}
\usepackage{threeparttable}
\usepackage{overpic}

\setlist[itemize]{leftmargin=*}
\setlist[enumerate]{leftmargin=*}
\setlist[description]{leftmargin=*}

\definecolor{midnightgreen}{rgb}{0.0, 0.29, 0.33}

\title{\textnormal{LongCat-Video-Avatar 1.6 Technical Report }}

\author{
    Meituan LongCat Team
}

\renewcommand{\headeright}{\raisebox{-0.2\height}{\includegraphics[height=1.8em]{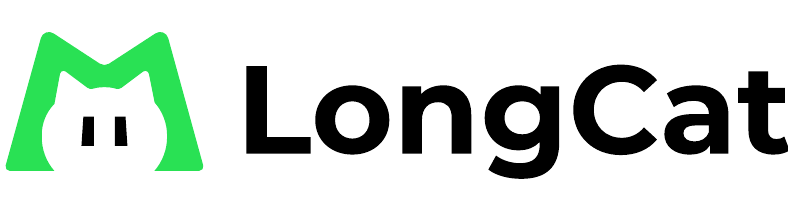}}}
\renewcommand{\shorttitle}{LongCat-Video-Avatar 1.6 Technical Report}

\usepackage{amsmath}
\begin{document}
\maketitle

\begin{abstract}

Recent advances in audio-driven avatar video generation have enabled increasingly realistic synthesis, yet achieving precise lip–audio synchronization, natural, human-like motion, and visual consistency over long sequences remains challenging. We present LongCat-Video-Avatar 1.6, which builds on LongCat-Video-Avatar 1.5 to improve expressive naturalness and reduce error accumulation during extended generation, \textit{e.g.,} exaggerated lip movements and overly red, temporally unstable lip coloration, enhance the naturalness of facial expressions and coordinated head and body movements.
These improvements are driven by a refined post-training pipeline that integrates reinforcement learning, improved classifier-free guidance strategies for DMD-based few-step distillation, long-video rollouts, and on-policy distillation. Together, these refinements improve expressive realism and enable the generation of videos exceeding 10 minutes while maintaining visual quality and temporal consistency. Arena evaluations show that LongCat-Video-Avatar 1.6 outperforms its predecessor and competing commercial audio-driven avatar models, demonstrating the effectiveness of the improved post-training pipeline.

\vspace{0.1cm}

\textbf{Page}: \href{https://meigen-ai.github.io/LongCat-Video-Avatar-1.6-Page}{https://meigen-ai.github.io/LongCat-Video-Avatar-1.6-Page} \\
\textbf{GitHub}: \href{https://github.com/meituan-longcat/LongCat-Video}{https://github.com/meituan-longcat/LongCat-Video}

  
  
  
\begin{figure}[htbp]
    \centering
    \includegraphics[width=0.75\textwidth]
        {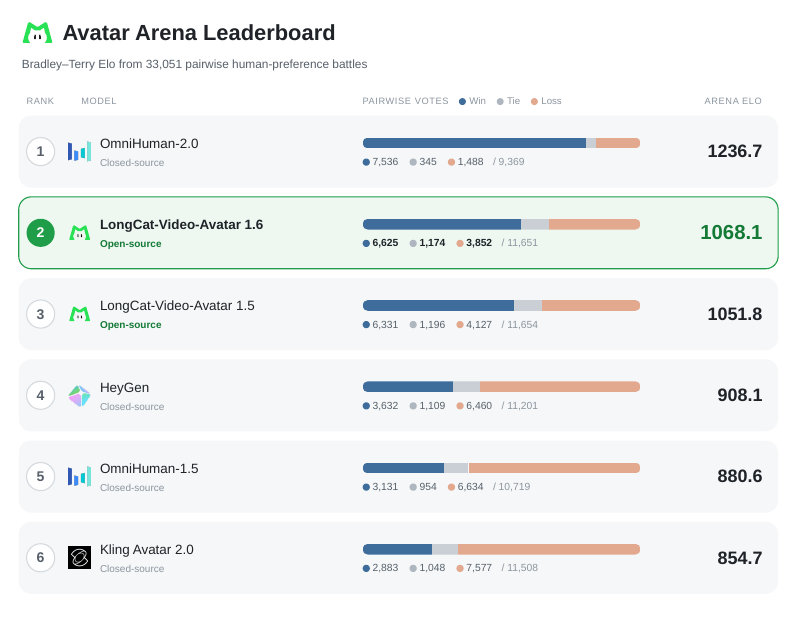}
\caption{
    \textbf{Arena.}
    Human preference comparison across six avatar generation models, based on 33,051 valid individual votes across 15 model pairs.
    Models are ranked by Bradley--Terry Elo scores, with higher scores indicating stronger human preference.
    The colored bars summarize each model's results against its five opponents.
    The counts beneath each bar report wins, ties, and losses, followed by the total number of votes involving that model.
    Open-source and closed-source labels indicate model access.
}
    \label{fig:avatar16_arena}
\end{figure}

\end{abstract}

\clearpage
\tableofcontents
\clearpage

\section{Introduction}
Audio-driven avatar video generation has advanced rapidly alongside large-scale video diffusion models~\cite{wang2025wan,team2025longcat}.
Recent portrait and human animation methods model speech-conditioned facial dynamics and body motion~\cite{wang2025fantasytalking,gan2025omniavatar}.
Multimodal conditioning further supports expressive character actions and flexible animation tasks~\cite{wei2025mocha,meng2026echomimicv3}.
Recent systems extend these capabilities to detailed human appearances and broader speech-conditioned motion, bringing digital humans closer to everyday conversational use~\cite{lin2025omnihuman,team2026longcat,semenov2026kandinsky}.
Research increasingly pursues perceptually natural articulation and movement~\cite{wang2026vibeavatar}, together with efficient generation over longer durations~\cite{li2026avatar, huang2026flowact}.
As these systems move from short demonstrations to sustained interaction, the quality of the speaking face becomes especially important: viewers continuously observe both how the mouth articulates speech and how the face changes over time.

Two limitations remain particularly visible in this setting.
First, matching the timing of speech does not necessarily produce natural articulation.
Recent lip-synchronization models target accurate audio--visual alignment and visual fidelity~\cite{daghigh2026highsync,cho2026lip}, while a natural speaking face also requires appropriate movement intensity and transitions between mouth shapes.
Mouth shapes and lip movements can appear exaggerated, tense, or strained even when they broadly follow the input audio.
A convincing avatar must preserve the distinctions needed for speech while producing relaxed movements and fluent transitions between mouth shapes.
Second, an initially plausible face can gradually deteriorate during long-video generation.
Long-duration portrait animation and video continuation have therefore become dedicated research targets~\cite{cui2025hallo2,yang2025infinitetalk,tu2025stableavatar}.
Streaming avatar systems also use explicit visual memory to preserve consistency over long histories~\cite{song2026interactiveavatar}.
In autoregressive continuation, small errors can enter the generated history and influence subsequent segments, allowing color shifts, texture changes, and local distortions to accumulate; progressively reddening lips are one conspicuous example.
Such degradation undermines visual consistency even when the character remains recognizable.
These problems affect different aspects of the same experience: natural speech-related motion within a segment and preservation of appearance across a prolonged conversation.

We therefore argue that a practical avatar should combine {natural mouth movements} with {resistance to long-term error accumulation}.
The first requirement concerns the quality of articulation as well as its synchronization with audio; reducing movement indiscriminately can suppress necessary speech dynamics, as the dynamic-collapse problem studied by DynaForcing~\cite{huang2026dynaforcing} illustrates.
The second requires controlling how errors propagate through successive continuations, beyond producing a visually clean opening clip; recent work on long-context video generation also emphasizes learning to continue from imperfect histories~\cite{chen2026context}.
Both must be achieved within an efficient end-to-end inference pipeline so that improved quality remains practical for long-duration use.
Recent few-step autoregressive video and lip-synchronization models reduce sampling costs~\cite{yin2025causvid,cho2026lip}, while streaming avatar systems combine acceleration with strategies for limiting long-horizon drift~\cite{huang2025live,li2025joyavatar}.

Post-training offers a direct way to refine these properties in an already capable avatar model.
Reward-guided post-training enables direct optimization of perceptual objectives in flow-matching models and audio-driven portrait generators~\cite{liu2026flow,tan2026flowportrait}.
In joint audio--video generation, preference-guided distillation also incorporates visual fidelity, speech naturalness, and synchronization rewards to preserve quality during acceleration~\cite{li2026hallolive}.
Existing work shows that perceptual rewards can improve generated avatar motion~\cite{wang2026vibeavatar}, while training on self-generated histories can reduce the mismatch between training and autoregressive inference~\cite{huang2026self,li2026avatar}.
We identify a complementary need for these two forms of supervision: relaxed, speech-appropriate articulation requires perceptual feedback on generated motion, whereas resistance to accumulated distortion requires learning from the histories that the model actually produces.
Clip-level optimization alone does not expose the model to the consequences of repeatedly conditioning on its own errors.
This motivates separate perceptual and long-video post-training branches whose updates can be combined during inference.

We present \textit{LongCat-Video-Avatar 1.6}, an open-source audio-driven avatar video generation framework focused on more natural speech-related mouth movements, reduced error accumulation in long videos, and efficient end-to-end inference.
Our {post-training framework} starts from a shared frozen base model.
The reinforcement learning branch uses Group Relative Policy Optimization (GRPO) to refine perceptual motion quality, with speech-related facial articulation as a central target.
The distillation branch first learns few-step generation and then performs long-video rollouts with on-policy self-distillation (OPSD) to improve continuation under model-generated histories.
Each branch learns its own LoRA adapter, enabling their updates to be combined in a single generation pipeline.

Within the distillation branch, we address the quality of individual segments before optimizing their repeated continuation.
DMD2-based distillation~\cite{yin2024improved} reduces sampling from 50 to eight denoising steps, providing an efficient generator for rollout training and inference.
We also revise the text--audio classifier-free guidance path to alleviate facial over-saturation, including an unnatural reddish tint around the lips.
Long-video rollout training then exposes the generator to its own preceding outputs, so that the supervision addresses errors after they have entered the continuation history.
For OPSD, we adapt the context-replacement strategy of OPSD-V~\cite{liu2026opsd} to our clip-level continuation.
An exponential moving average (EMA) teacher provides velocity targets at the denoising states visited by the student.
For continuation clips, the teacher constructs its context from real video frames and one generated boundary frame from its previous prediction.
This combines supervision grounded in real video with exposure to the student's own rollout errors, targeting more reliable continuation from imperfect histories.
Together, perceptual refinement, guidance adjustment, and rollout-based learning target natural articulation and sustained visual consistency at a practical sampling cost.

\begin{figure}[p]
  \centering
    \includegraphics[width=1.0\linewidth]{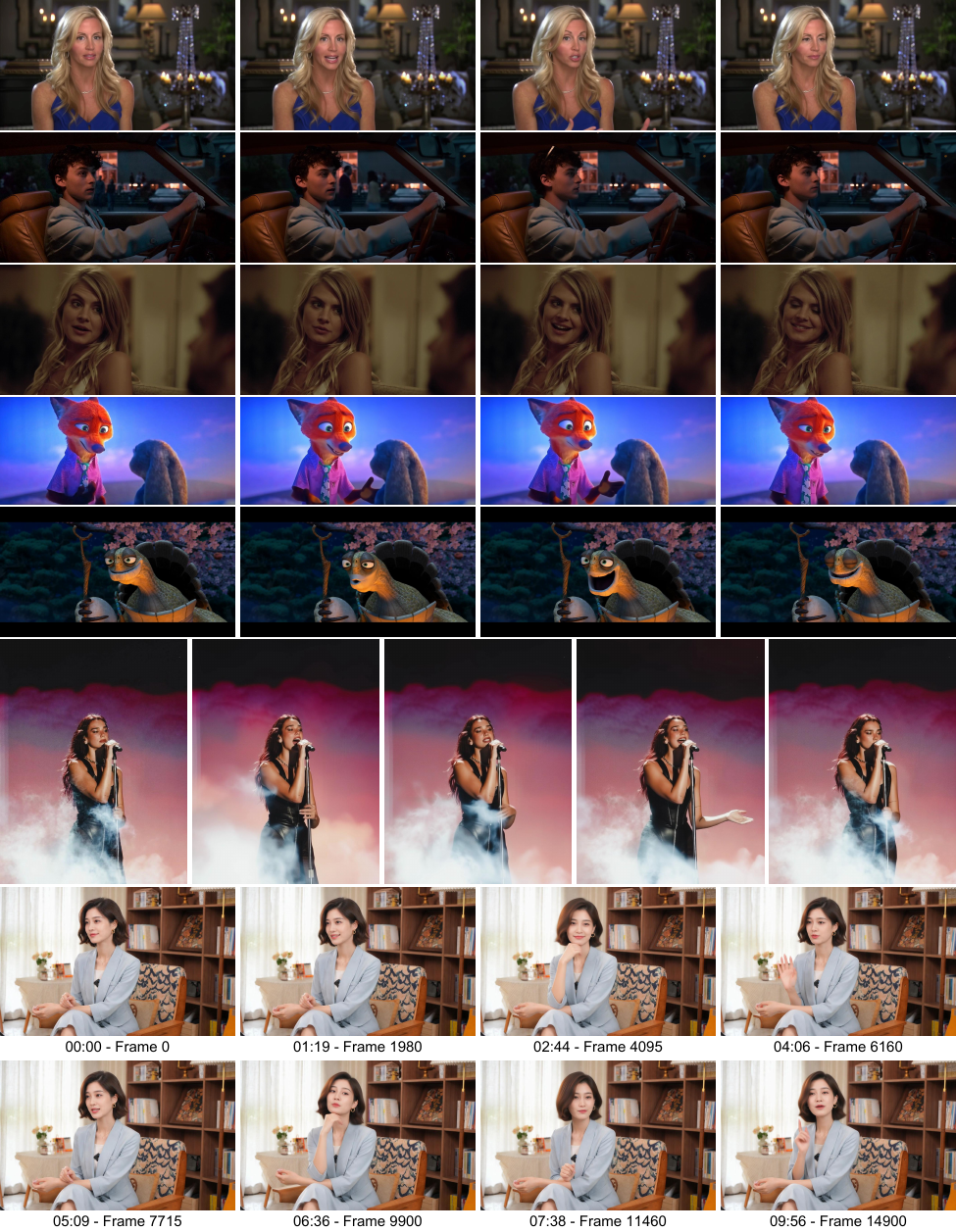}
      \caption{Demonstration of video frames generated by LongCat-Video-Avatar 1.6 across various application scenarios, including interviews, acting, singing, and animation.}
      \label{fig:ours-qualitative-showcase}
\end{figure}

As shown in Figure~\ref{fig:avatar16_arena}, LongCat-Video-Avatar 1.6 ranks second among six evaluated systems in a human preference Arena comprising 33,051 valid individual votes.
With a Bradley--Terry Elo score of 1068.1, it ranks behind OmniHuman-2.0 and ahead of LongCat-Video-Avatar 1.5, HeyGen, OmniHuman-1.5, and Kling Avatar 2.0.
As discussed in Section~\ref{sec:eval-arena}, this preference gap may partly reflect OmniHuman-2.0's apparent enhancement of reference detail, while our model more closely preserves the input appearance.

The main contributions of this report are summarized as follows:
\begin{itemize}
    \item We develop a post-training framework targeting relaxed, speech-appropriate mouth movements and reduced accumulation of visual errors during long-video avatar generation.
    \item We combine perceptual optimization with eight-step distillation and a revised text--audio guidance path to support natural articulation, facial color fidelity, and efficient generation.
    \item We introduce long-video rollout training with on-policy self-distillation, using additional teacher context to guide the student on its own generated histories and improve appearance preservation during continuation.
\end{itemize}

\section{Method}
\subsection{Post-training Pipeline}

\begin{figure}
    \centering
    \includegraphics[width=1.0\linewidth]{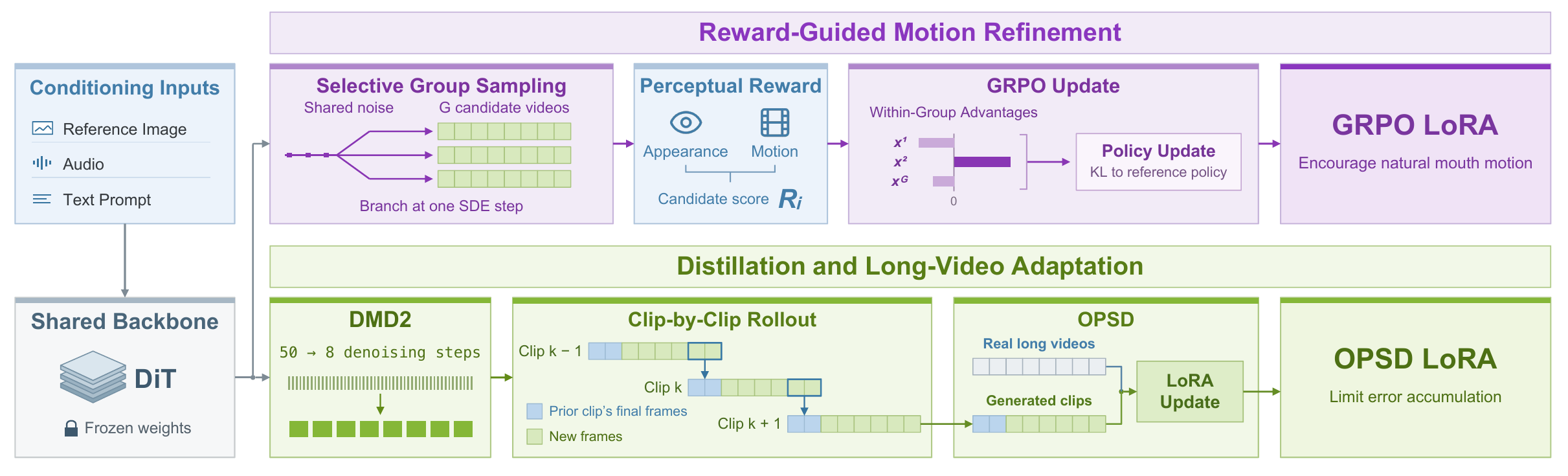}
    \caption{Overview of our post-training framework. The GRPO branch performs reward-guided reinforcement learning, while the DMD--OPSD branch first conducts clip-level few-step distillation and then applies long-video on-policy self-distillation. The resulting GRPO and OPSD LoRA adapters are jointly incorporated into the frozen base model during inference.}
    \label{fig:post_train_pipe}
\end{figure}

As illustrated in Fig.~\ref{fig:post_train_pipe}, our post-training pipeline consists of two complementary branches: reinforcement learning with Group Relative Policy Optimization (GRPO) and few-step distillation followed by on-policy self-distillation. Both branches start from a well-trained audio-driven video generation model, whose training pipeline follows LongCat-Video-Avatar~1.5 \cite{team2026longcat}. During post-training, the base model remains frozen, while each branch independently optimizes a lightweight LoRA adapter. The resulting adapters are jointly activated during inference, allowing the model to benefit from both reward-guided optimization and efficient long-video generation without modifying the base model parameters.

\subsection{Reinforcement Learning}

Following the GRPO strategy in LongCat-Video~\cite{team2025longcat}, we optimize the audio-driven generator with reward-guided updates to improve the naturalness of facial expressions and body movements. For each conditioning input, consisting of a reference image, an audio segment, and a text prompt, the policy generates a group of candidate videos. Their rewards are compared within the group to estimate relative advantages, allowing the model to favor higher-reward outputs without training a separate value network. Only the GRPO LoRA adapter is updated, while the pretrained backbone remains frozen.

For flow-matching generation, we follow the selective stochastic sampling scheme of LongCat-Video. Candidates in the same group share the initial noise and a randomly selected denoising timestep. Independent noise is injected through a stochastic differential equation (SDE) step only at this timestep; all remaining steps use deterministic ordinary differential equation (ODE) sampling. This isolates the source of variation between candidates and makes the final reward more directly attributable to the selected transition.

Let $c_j$ denote the conditioning input for group $j$, and let $\{\hat{x}_{0,j}^{i}\}_{i=1}^{G}$ be its $G$ generated videos. Given a scalar reward $R$, the relative advantage is
\begin{equation}
\hat{A}_{j}^{i}=\frac{R(\hat{x}_{0,j}^{i},c_j)-\mu_j}{\sigma_{\max}+\epsilon},
\qquad
\mu_j=\frac{1}{G}\sum_{i=1}^{G}R(\hat{x}_{0,j}^{i},c_j).
\label{eq:grpo_advantage}
\end{equation}
Here, $\sigma_{\max}$ is the maximum within-group reward standard deviation across the batch, and $\epsilon>0$ ensures numerical stability. This normalization reduces the influence of groups with weak reward differences.

At the selected stochastic timestep $t$, we maximize the reweighted GRPO objective
\begin{equation}
\mathcal{J}_{\mathrm{GRPO}}(\theta)
=\mathbb{E}\!\left[\frac{1}{G}\sum_{i=1}^{G}\left(
\lambda_{\mathrm{policy}}\rho_j^i(\theta)\hat{A}_j^i
-\beta\lambda_{\mathrm{KL}}D_{\mathrm{KL}}\!\left(p_\theta^{j,i}\,\|\,p_{\mathrm{ref}}^{j,i}\right)
\right)\right].
\label{eq:grpo_objective}
\end{equation}
The expectation is over conditioning inputs, selected timesteps, and trajectories sampled from the old policy. Here, $p_\theta^{j,i}=\pi_\theta(\cdot\mid z_t^{j,i},c_j)$ denotes the stochastic transition distribution, and $\rho_j^i(\theta)=p_\theta^{j,i}(z_{t-\Delta t}^{j,i})/p_{\mathrm{old}}^{j,i}(z_{t-\Delta t}^{j,i})$ is its likelihood ratio relative to the sampling policy. The frozen reference policy defines $p_{\mathrm{ref}}^{j,i}$, and $\beta$ controls the KL penalty. Following LongCat-Video, the reweighting factors are $\lambda_{\mathrm{policy}}=\sqrt{t/[\Delta t(1-t)]}$ and $\lambda_{\mathrm{KL}}=t/[\Delta t(1-t)]$, where $t\in(0,1)$ is the normalized noise level and $\Delta t>0$ is the step size. They compensate for variation in gradient scale across the sampling schedule. Only the GRPO LoRA parameters $\theta$ are optimized, and the resulting adapter is combined with the OPSD adapter at inference.

\subsection{Distillation Learning}
\subsubsection{Few-step Distillation}
Inspired by Distribution Matching Distillation 2 (DMD2)~\cite{yin2024improved}, we distill the multi-step video diffusion model into an efficient few-step generator. DMD2 aligns the distribution of the generator with that of the teacher model by minimizing the reverse Kullback--Leibler (KL) divergence. Its standard implementation, however, incurs substantial GPU memory overhead because three model instances with the same architecture—the generator, fake-score estimator, and real-score estimator—must be maintained simultaneously.

To reduce this memory overhead, we introduce a parameter-efficient architecture based on a single shared Diffusion Transformer (DiT) backbone with multiple LoRA adapters. Specifically, the backbone dynamically switches between a Generator LoRA and a Fake-Score LoRA to perform few-step generation and fake-score estimation, respectively. The frozen base DiT serves as the real-score estimator and provides guidance from the original data distribution. This design avoids loading three complete DiT models while retaining the distribution-matching objective of DMD2.

To balance generation quality and inference efficiency, we distill the model from 50-step inference to eight denoising steps. Both the real- and fake-score estimators inherit the timestep scheduler from the preceding training stage to maintain a consistent noise scale. We observe that excessively long training gradually degrades visual realism. Therefore, we stop distillation after 600 optimization steps, at which point the model achieves its best generation fidelity before the onset of quality degradation. The learning rates of the generator and fake-score estimator are set to $2\times10^{-5}$ and $4\times10^{-6}$, respectively, with a generator-to-fake-scorer update ratio of $1{:}5$.

\begin{figure}
    \centering
    \includegraphics[width=1.0\linewidth]{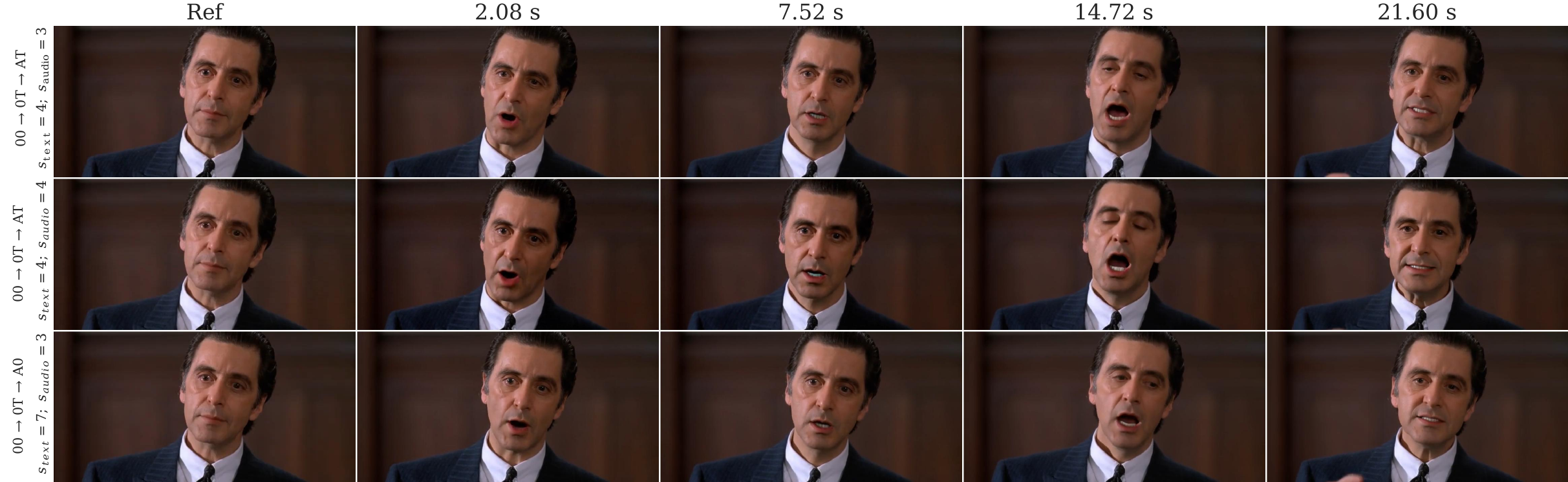}
    \caption{Qualitative comparison of different classifier-free guidance (CFG) paths and guidance strengths. The columns show the reference image and generated frames at different timestamps.}
    \label{fig:cfg_path_comp}
\end{figure}

We further find that the classifier-free guidance (CFG) strategy has a notable impact on visual quality and color stability. A uniform guidance configuration tends to produce over-saturated facial regions, particularly an unnatural reddish tint around the lips. Adjusting the condition-guidance path effectively alleviates this artifact and improves the overall visual appearance. Based on our empirical evaluation, we adopt the guidance path
$\mathrm{00}\!\rightarrow\!\mathrm{0T}\!\rightarrow\!\mathrm{A0}$:
\begin{equation}
\begin{aligned}
v_{\theta}
={}&v_{00}
+s_{\mathrm{text}}
    \left(v_{0T}-v_{00}\right) 
+s_{\mathrm{audio}}
    \left(v_{A0}-v_{0T}\right), 
\end{aligned}
\label{eq:audio_text_cfg}
\end{equation}
where $v_{00}$, $v_{0T}$, and $v_{A0}$ denote the unconditional, text-conditioned, and audio-conditioned velocity fields predicted by the model, respectively, and $v_{\theta}$ is the guided velocity field. We observe that visual quality is sensitive to the audio-guidance strength. Increasing the audio CFG scale to $s_{\mathrm{audio}}=4$ introduces noticeable color artifacts, including overly reddish lips and a greenish tint on the teeth, as shown in Fig.~\ref{fig:cfg_path_comp}. Reducing the audio-guidance scale to $3$ visibly alleviates these artifacts. Moreover, among the evaluated configurations, the guidance path
$\mathrm{00}\!\rightarrow\!\mathrm{0T}\!\rightarrow\!\mathrm{A0}$
produces a more natural facial appearance than
$\mathrm{00}\!\rightarrow\!\mathrm{0T}\!\rightarrow\!\mathrm{AT}$.
Based on these observations, we adopt
$\mathrm{00}\!\rightarrow\!\mathrm{0T}\!\rightarrow\!\mathrm{A0}$
with $s_{\mathrm{text}}=7$ and $s_{\mathrm{audio}}=3$ as our final CFG configuration.

\subsubsection{Rollout for Long Video Generation}
\label{sec:rollout}

Existing methods for long video generation follow two main paradigms. The first turns a video diffusion model into a causal generator, as in CausVid~\cite{yin2025causvid} and Self Forcing~\cite{huang2026self}. A block-wise causal attention mask restricts each chunk to attend only to itself and earlier chunks, and the video is generated chunk by chunk with a KV cache. The second paradigm, adopted by InfiniteTalk~\cite{yang2025infinitetalk} and LongCat-Video~\cite{team2025longcat}, keeps the model fully bidirectional within each clip and extends the video by repeatedly continuing from the last frames of the previously generated clip. We refer to this paradigm as autoregressive video continuation: generation is autoregressive across clips, while each clip is still denoised with full bidirectional attention. LongCat-Video-Avatar follows the second paradigm.

Formally, let $F$ and $C$ denote the number of video frames in a clip and its overlap context, respectively, and let $L$ and $L_{\mathrm{ctx}}$ denote the corresponding numbers of latent frames after VAE encoding. The stride between clips is $S=F-C$. In a rollout of $N$ clips, clip $k\in\{0,\ldots,N-1\}$ covers frames $[kS,\,kS+F)$ and is driven by the audio window $\mathbf{A}_k$ over the same interval. Let $\mathcal{E}$ and $\mathcal{D}$ denote the VAE encoder and decoder, $x_{\mathrm{ref}}$ the reference frame, and $c_{\mathrm{text}}$ the text prompt. The first clip is generated in the image-audio-to-video (IA2V) setting, with the reference latent $z_{\mathrm{ref}}=\mathcal{E}(x_{\mathrm{ref}})$ placed at the first temporal position. For $k\ge1$, we decode the generated latents $\hat{z}_{k-1}$ of the previous clip to obtain the video $\hat{x}_{k-1}=\mathcal{D}(\hat{z}_{k-1})$. We jointly re-encode its last $C$ frames to obtain the context latents of clip $k$.
\begin{equation}
z_{\mathrm{ctx},k}=\mathcal{E}\!\left(\hat{x}_{k-1}[F-C:F]\right).
\label{eq:rollout_context}
\end{equation}
The context latents $z_{\mathrm{ctx},k}$ occupy the first $L_{\mathrm{ctx}}$ temporal positions of clip $k$ and are concatenated with Gaussian noise along the temporal axis. The context latents are kept at timestep zero, and only the remaining $L-L_{\mathrm{ctx}}$ positions are denoised. The reference latent $z_{\mathrm{ref}}$ is additionally provided as an identity condition for continuation clips, so that the appearance stays anchored to $x_{\mathrm{ref}}$. When the clips are assembled, the first clip is kept in full, and for every subsequent clip only the $S$ newly generated frames are appended.

Each clip is sampled with the eight-step generator described above. Since the procedure has no fixed horizon, it can produce videos of arbitrary length. However, it also introduces a mismatch between training and inference. The few-step generator is distilled on individual clips, whose temporal contexts are derived from real videos, whereas during rollout every context after the first clip comes from the model's own imperfect outputs. Small deviations in color, texture, or the lip region are re-encoded into the next context and amplified over successive clips, which appears as progressive color drift and loss of detail in long videos. The next section addresses this mismatch by training directly on such rollouts.

\subsubsection{On-policy Self-Distillation}
\label{sec:opsd}

We adapt OPSD-V~\cite{liu2026opsd} to audio-driven avatar continuation by applying context replacement to the overlapping frames between clips. The student follows the eight-step rollout in Sec.~\ref{sec:rollout}. At each denoising step, the EMA teacher receives the same noisy latents and audio, text, and reference conditions as the student, with a modified temporal context.

For continuation clip $k\ge1$, the teacher context covers the same $C$ frames as the student context in Eq.~\eqref{eq:rollout_context}. We take the first $C-1$ frames from the corresponding interval of the real video $y$ and the final frame of the teacher-generated video $\hat{x}^{t}_{k-1}=\mathcal{D}(\hat{z}^{t}_{k-1})$ for the previous clip. We encode these frames jointly to obtain the teacher context latents.
\begin{equation}
z^{t}_{\mathrm{ctx},k}
=\mathcal{E}\!\left(\left[\,y[kS:kS+C-1],\;\hat{x}^{t}_{k-1}[F-1]\,\right]\right).
\label{eq:opsd_teacher_context}
\end{equation}
For the first clip, the student's temporal context is the reference latent $z_{\mathrm{ref}}$, and the teacher uses $z^{t}_{\mathrm{ctx},0}=\mathcal{E}([\,y[0:C-1],\,x_{\mathrm{ref}}\,])$.

Following OPSD-V, we match the student and teacher velocity fields $v^{s}_{k,j}$ and $v^{t}_{k,j}$ at the same student-visited state. The loss is computed on the latent suffix $[L_{\mathrm{ctx}},L)$ and averaged over all clips $k$ and denoising steps $j$.
\begin{equation}
\mathcal{L}_{\mathrm{OPSD}}=\mathbb{E}_{k,j}\left[\left\|\,v^{s}_{k,j}[L_{\mathrm{ctx}}\!:\!L]-\operatorname{sg}\!\left(v^{t}_{k,j}[L_{\mathrm{ctx}}\!:\!L]\right)\right\|_2^2\right],
\label{eq:opsd_loss}
\end{equation}
where $\operatorname{sg}$ denotes stop-gradient. At inference, we use the student OPSD adapter together with the GRPO adapter in the rollout of Sec.~\ref{sec:rollout}.

\section{Data}
\label{sec:data}

    LongCat-Video-Avatar 1.6 largely reuses the data resources and curation framework of LongCat-Video-Avatar 1.5~\cite{team2026longcat}.
The source pool includes close-up facial videos, interviews, acted performances, human--object interactions, music performances, and stylized content.
Shared annotations describe subject geometry, speaking activity, audio--visual correspondence, visual quality, camera motion, and semantic content.
From these resources, we prepare conditioning examples for perceptual optimization, aligned short clips for few-step distillation, and continuous video sequences for long-video self-distillation.

\subsection{Data for Reinforcement Learning}

For GRPO, each conditioning example comprises a reference image, an audio segment, and a text prompt.
Examples cover everyday conversation, expressive performance, and non-speaking behavior, including gaze changes, head motion, and gestures during pauses.
Face and body annotations describe the visible subjects and shot composition, while speaking-activity annotations associate subjects with the driving audio.
Captions describe facial, head, and body movements together with scene and interaction context, and are matched to the sampled temporal interval.
Speech-state and emotional-expression annotations further characterize pauses and expressive dynamics, providing context for evaluating speech-appropriate articulation and motion.

\subsection{Data for Few-Step Distillation}

Few-step distillation uses short clips with aligned visual, audio, and text conditions.
The audio pipeline extracts vocal tracks and filters samples with unreliable synchronization between speech and visible mouth motion.
Each sampled training window is checked for duration, frame rate, resolution, brightness, border artifacts, frame discontinuities, and abnormal motion.
Face visibility, body composition, and camera-motion annotations characterize the visual conditions.
The vocal track and local caption are matched to the same video interval, preserving the correspondence between speech, articulation, and visual content.

\subsection{Data for On-Policy Self-Distillation}

For on-policy self-distillation, we use real videos of people speaking, paired with aligned audio and video captions. The data include both face-cropped videos and videos retaining the original framing, providing facial details as well as broader visual context. We sample continuous sequences of approximately 20 seconds at a target frame rate of 25 FPS to support training over multiple consecutive clips.
Each sequence provides a reference image, audio, and visual captions as generation conditions. Real frames at corresponding temporal positions supply context for the teacher, while the student generates successive clips using its own outputs as history. This enables on-policy training with real-video guidance, helping reduce error accumulation and improve visual consistency during extended generation.

\begingroup
\raggedbottom
\setlength{\textfloatsep}{12pt plus 2pt minus 2pt}
\setlength{\floatsep}{10pt plus 2pt minus 2pt}
\setlength{\intextsep}{10pt plus 2pt minus 2pt}
\captionsetup{font=small,skip=5pt}
\renewcommand{\topfraction}{0.9}
\renewcommand{\textfraction}{0.08}

\begin{figure}
    \centering
    \includegraphics[width=\linewidth]{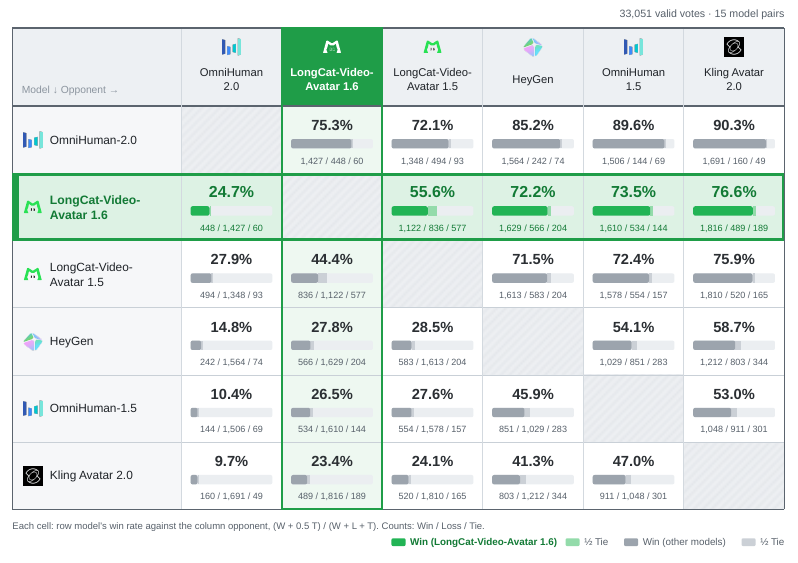}
    \caption{\textbf{Pairwise human preference across six avatar models.}
    Each off-diagonal cell shows the row model's win rate against the column opponent; the numbers below the bar give win / loss / tie counts.
    The win rates are calculated as Eq.~\ref{eq:adjusted_win_rate}.
    Results are based on 33,051 valid individual votes across 15 model pairs, with percentages rounded to one decimal place.
    Green highlights LongCat-Video-Avatar 1.6.}
    \label{fig:sota-preference}
\end{figure}

\section{Comparison with SOTA Commercial Models}
\label{sec:eval-sota}
\label{sec:eval-arena}

\subsection{Arena Settings}

We compare LongCat-Video-Avatar 1.6 with its predecessor, 
LongCat-Video-Avatar 1.5, and four commercial models: 
OmniHuman-2.0~\cite{byteplus2026omnihuman2}, HeyGen~\cite{HeyGen}, OmniHuman-1.5~\cite{jiang2025omnihuman}, and Kling Avatar 2.0~\cite{team2025klingavatar2}.
The evaluation set consists of 508 image–speech pairs from EvalTalker~\cite{zhou2025evaltalker}, and each model is tasked with generating a video for every pair. 
For each evaluation task, two available outputs produced by different models from the same input are randomly selected, presented side by side, and assigned to an annotator for comparison. 
The evaluation covers all 15 possible pairings among the six models and yields 33{,}051 valid individual votes. 
Due to occasional API failures, some model--input combinations did not produce valid videos. 
Consequently, the number of evaluated samples varies across model pairings.

Before participating in the formal evaluation, annotators are required to complete a qualification task. 
Each qualification item is reviewed by multiple experts and retained only when they agree that it has a single, unambiguous correct answer. Annotators must achieve an accuracy of at least 80\% on the qualification task to proceed to the formal evaluation.
Each formal evaluation task consists of one single-choice question and one multiple-choice question. 
For the single-choice question, annotators select one of five options: ``Video 1 is better,'' ``Video 2 is better,'' ``Both are good,'' ``Both are bad,'' or ``The video cannot be played.'' 
For the multiple-choice question, annotators may select one or more reasons for their preference, including greater naturalness and coordination, better physical plausibility, better audio--visual synchronization, higher visual quality, greater temporal stability, and other reasons. Annotators may additionally provide free-text comments to explain their judgments.

The responses ``Both are good'' and ``Both are bad'' are treated as ties, with each model receiving half a win. Responses indicating that a video cannot be played are excluded from the valid preference votes. The tie-adjusted pairwise win rate is defined as
\begin{equation}
    \operatorname{WinRate}_{\mathrm{adjusted}}
    =
    \frac{W + 0.5T}{W + L + T},
    \label{eq:adjusted_win_rate}
\end{equation}
where $W$, $L$, and $T$ denote the numbers of wins, losses, and ties, respectively. We further analyze the selected preference reasons together with the supplementary free-text comments to characterize the relative strengths and weaknesses of each model.


\subsection{Results}

\paragraph{Overall ranking and pairwise preference.}
Figure~\ref{fig:avatar16_arena} summarizes the overall Arena ranking.
LongCat-Video-Avatar 1.6 ranks second among the six evaluated models with a Bradley--Terry Elo score of 1068.1, behind OmniHuman-2.0 (1236.7) and ahead of LongCat-Video-Avatar 1.5 (1051.8), HeyGen (908.1), OmniHuman-1.5 (880.6), and Kling Avatar 2.0 (854.7).
Figure~\ref{fig:sota-preference} provides the detailed head-to-head results.
LongCat-Video-Avatar 1.6 receives 72.2\%, 73.5\%, and 76.6\% of valid votes against HeyGen, OmniHuman-1.5, and Kling Avatar 2.0, respectively.
In the direct comparison with LongCat-Video-Avatar 1.5, LongCat-Video-Avatar 1.6 receives 55.6\% of the 2,535 valid votes, compared with 44.4\% for LongCat-Video-Avatar 1.5.
A gap remains to OmniHuman-2.0: it wins 75.3\% of their direct comparisons, versus 24.7\% for LongCat-Video-Avatar 1.6.


\begin{figure}
    \centering
    \includegraphics[width=\linewidth]{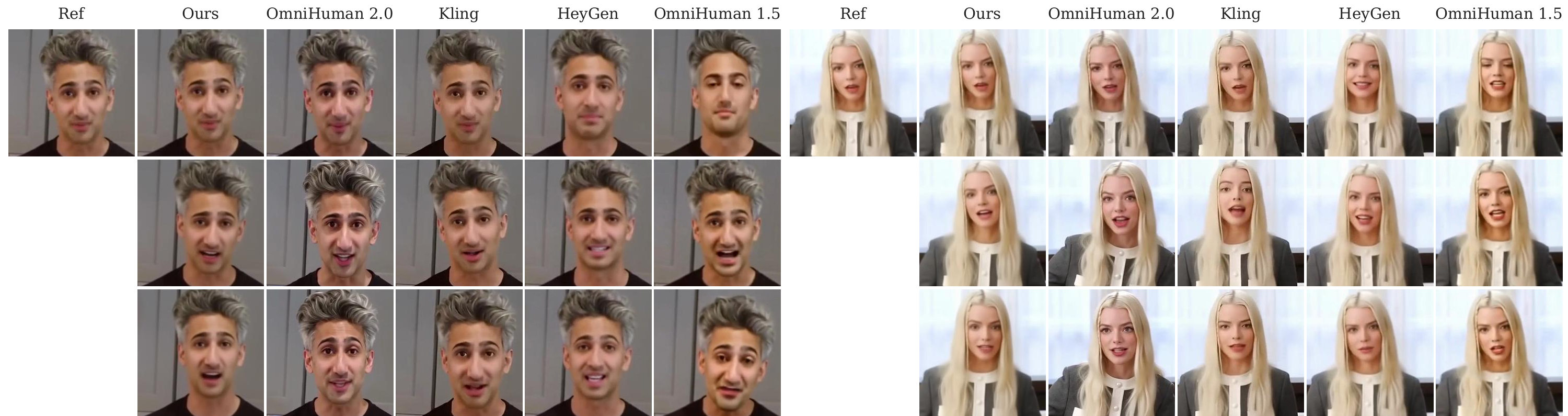}
    \caption{\textbf{Reference appearance and generated facial detail.}
    Within each example, the reference crop occupies the upper-left cell, followed by columns for LongCat-Video-Avatar 1.6 (Ours), OmniHuman-2.0, Kling Avatar 2.0, HeyGen, and OmniHuman-1.5.
    The three rows show chronologically ordered generated frames, beginning with the first frame.
    OmniHuman-2.0 shows increased apparent detail relative to the reference, while Ours more closely preserves its original appearance.}
    \label{fig:reference-appearance}
\end{figure}

\begin{figure}
    \centering
    \includegraphics[width=\linewidth]{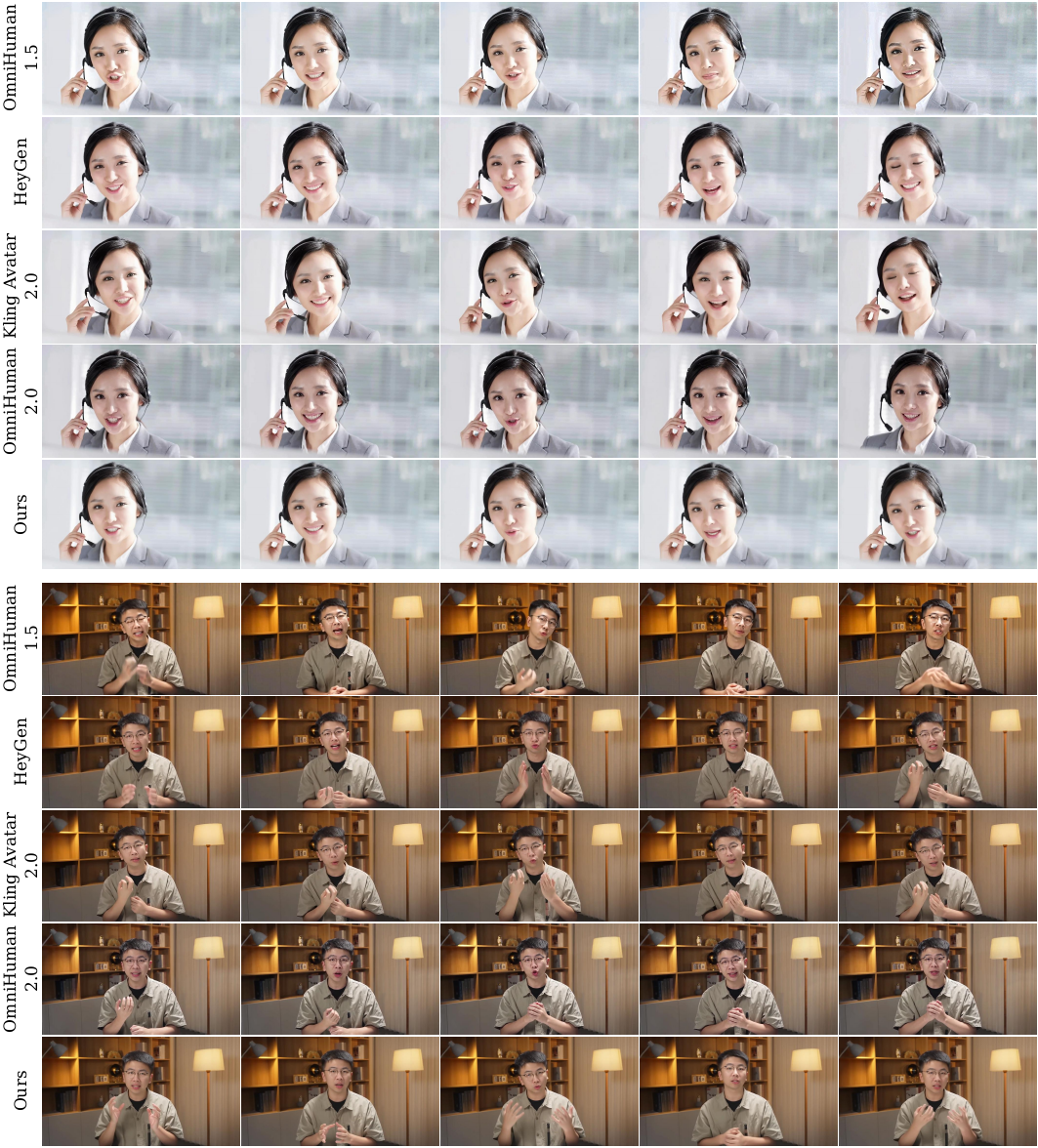}
    \caption{\textbf{Qualitative comparison with commercial avatar systems.}
    Two examples are stacked vertically, and within each example, rows show OmniHuman-1.5, HeyGen, Kling Avatar 2.0, OmniHuman-2.0, and LongCat-Video-Avatar 1.6 (Ours), while columns show five chronologically ordered frames aligned by the source-video timeline.}
    \label{fig:qualitative-sota}
\end{figure}

\begin{figure}
    \centering
    \includegraphics[width=0.8\linewidth]{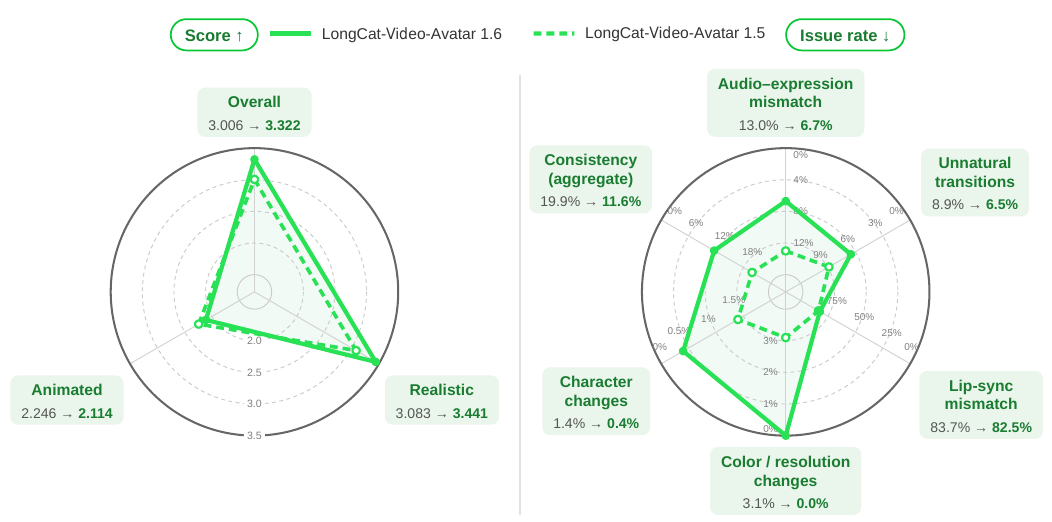}
    \caption{\textbf{Comparison of LongCat-Video-Avatar 1.5 and 1.6.}
    Solid and dashed green curves denote versions 1.6 and 1.5, respectively. 
    Left: Human-likeness ratings on a five-point scale, where higher scores are better. Right: Expert-annotated issue rates, with lower values indicating better performance. The issue-rate axes use independent, reversed radial scales, with zero located at the outer boundary.}
    \label{fig:version-improvements}
\end{figure}

\paragraph{Analysis of human preferences.}
Naturalness and audio--visual correspondence are recurring reasons for preferring LongCat-Video-Avatar 1.6 over its predecessor, HeyGen, OmniHuman-1.5, and Kling Avatar 2.0.
The comparison with OmniHuman-2.0 also highlights the role of image rendering in overall preference: greater apparent detail can improve perceived visual quality.
Figure~\ref{fig:reference-appearance} compares the reference images with three generated frames from each system.
In these examples, OmniHuman-2.0 appears to enhance the reference appearance, producing stronger local contrast and more pronounced facial and hair detail.
This change is visible in the early frames and remains evident in subsequent frames, where its output appears sharper than both the reference and LongCat-Video-Avatar 1.6.
Our model more closely preserves the reference image's original appearance and softer texture rather than introducing a comparable increase in apparent detail.
This difference may contribute to the visual-quality advantage of OmniHuman-2.0 and help explain its higher overall preference in the Arena.
Appearance enhancement and natural expressive behavior are nevertheless distinct aspects of quality.
The qualitative examples in Figure~\ref{fig:qualitative-sota} show that OmniHuman-2.0 can also produce exaggerated mouth shapes and expressions, while LongCat-Video-Avatar 1.6 maintains more restrained articulation.

\paragraph{Qualitative comparison.}
Figure~\ref{fig:qualitative-sota} compares the five systems on two selected examples, with five temporally aligned frames per method.
In the upper example, OmniHuman-1.5 and OmniHuman-2.0 exhibit exaggerated mouth shapes at selected speech moments, while Kling Avatar 2.0 shows an irregular mouth contour and HeyGen shows asymmetry between the upper and lower lips.
LongCat-Video-Avatar 1.6 maintains comparatively restrained articulation across the displayed frames.
The later frames also reveal a change in facial appearance in OmniHuman-1.5.
In the lower example, OmniHuman-1.5, HeyGen, and OmniHuman-2.0 show exaggerated mouth opening or lip protrusion at the selected speech moment, whereas LongCat-Video-Avatar 1.6 produces a more moderate mouth shape.
The selected frames also show unnatural hand poses in HeyGen and a mismatch between facial expression and hand posture in Kling Avatar 2.0; OmniHuman-1.5 develops an increasingly mannered appearance across the sequence.
These cases complement the appearance comparison in Figure~\ref{fig:reference-appearance}, illustrating that sharper facial rendering can coexist with exaggerated articulation.
\par


\section{Comparison with the Predecessor }
\label{sec:evaluation}



\begin{figure}[!t]
    \centering
    \includegraphics[width=\linewidth]{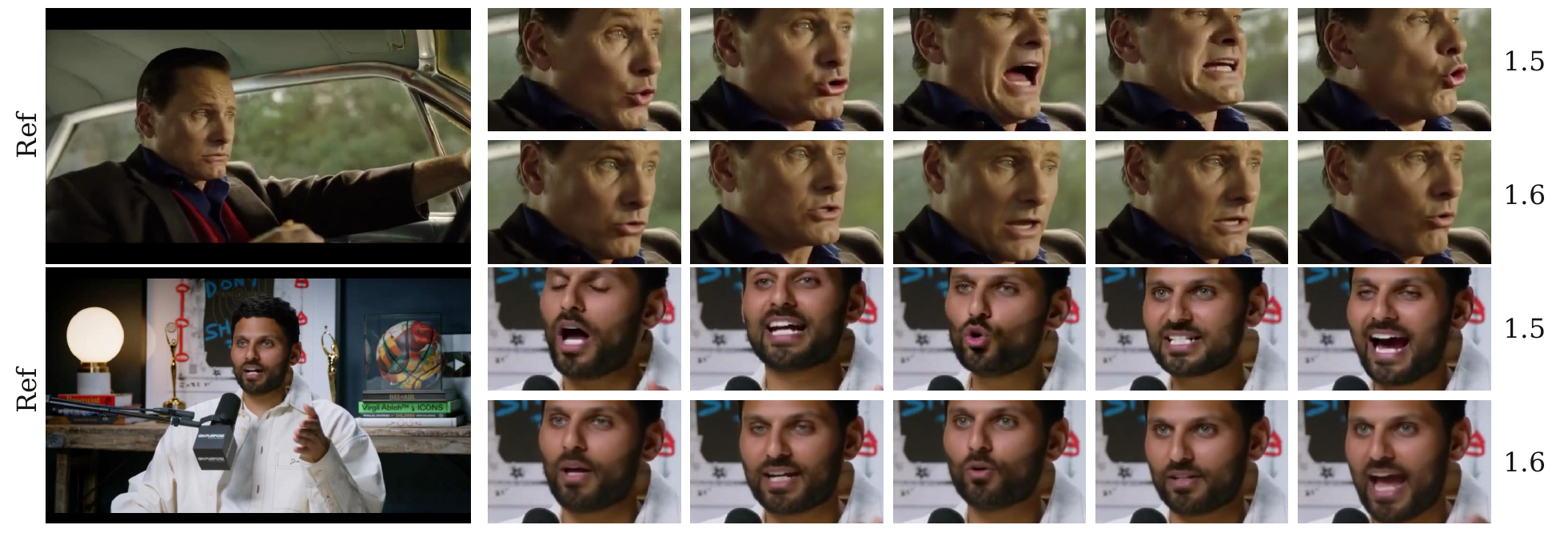}
    \caption{\textbf{Mouth naturalness in LongCat-Video-Avatar 1.5 and 1.6.}
    Two examples are stacked vertically, each showing the initial frame on the left and enlarged facial crops from selected generated frames on the right. 
    Within each example, the upper and lower rows correspond to 1.5 and 1.6, respectively. 
    In the selected frames, 1.6 shows smaller mouth openings and less pronounced lip protrusion and stretching.}
    \label{fig:mouth-naturalness}

    \vspace{8pt}
    \centering
    \includegraphics[width=\linewidth]{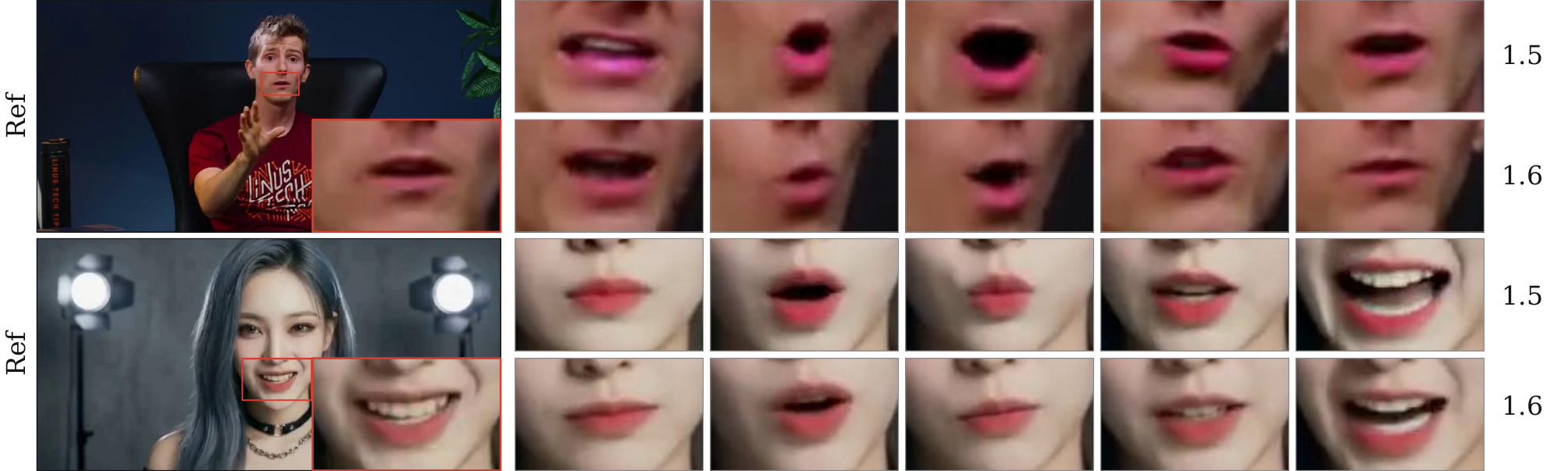}
    \caption{\textbf{Lip-color preservation during continued generation.}
    Two examples are stacked vertically.
    Each example shows the initial frame and an enlarged mouth inset on the left, followed by mouth crops from selected generated frames on the right.
    The upper and lower rows correspond to LongCat-Video-Avatar 1.5 and 1.6, respectively.
    1.6 remains closer to the initial lip tone, while 1.5 shows visible color shifts and over-saturation.}
    \label{fig:lip-color-consistency}
    \label{fig:appearance-over-time}
\end{figure}

\subsection{Settings}
\label{sec:eval-settings}

\paragraph{Benchmark.}
Our human-likeness benchmark builds on EvalTalker~\cite{zhou2025evaltalker} and comprises 508 image--audio pairs.
Figure~\ref{fig:version-improvements} reports human-likeness scores for the full benchmark (Overall) and the Realistic and Animated subsets.
The accompanying expert assessment reports six issue measures: audio--expression mismatch, unnatural expression transitions, lip-synchronization mismatch, color or resolution changes, character changes, and aggregate consistency issues.


\paragraph{Human-likeness ratings.}
The absolute-rating study involves 270 participants and follows a three-annotation protocol for each generated video.
Participants assign a score from 1 to 5, with higher scores indicating more human-like outputs.
After filtering invalid annotations, we report the overall mean and the means for realistic and animated inputs.




\subsection{Results}
\label{sec:eval-version}

Figure~\ref{fig:version-improvements} compares LongCat-Video-Avatar 1.5 and 1.6 in two groups: human-likeness scores on the left and expert issue rates on the right.
The overall human-likeness score increases from 3.006 to 3.322.
The mean score for realistic inputs increases from 3.083 to 3.441, while that for animated inputs decreases from 2.246 to 2.114.
Long-duration animated videos are relatively scarce in our OPSD training set. 
This limited coverage may contribute to slightly reduced generalization to animated inputs after post-training.
The following paragraphs examine the speech-related and appearance-related issue rates alongside qualitative examples of mouth naturalness and appearance preservation.


\paragraph{Mouth Naturalness and Speech Synchronization}
\label{sec:eval-mouth}

Natural articulation requires mouth shapes and movement amplitudes appropriate to the speech content and expressive intensity.
Figure~\ref{fig:mouth-naturalness} compares the two versions at selected speech moments.
LongCat-Video-Avatar 1.5 exhibits exaggerated mouth openings, pronounced lip protrusion, and visibly strained lip stretching.
LongCat-Video-Avatar 1.6 produces more relaxed mouth shapes while retaining distinct speech-related articulation.
The upper example particularly illustrates the reduction in excessive jaw opening and tense lip contours.
\par

For speech-related expression dynamics, 
the issue-rate group in Figure~\ref{fig:version-improvements} shows that expression/action mismatch with audio decreases from 13.0\% to 6.7\%, and unnatural expression transitions decrease from 8.9\% to 6.5\%.
These results indicate improved correspondence between speech and expression, together with more natural expression transitions.
Meanwhile, the overall lip-synchronization mismatch rate decreases from 83.7\% to 82.5\%.
Notice that the lip-sync mismatch rate is checked by playing the video at 0.5x speed and this rigorous review process results in a relatively high error rate.

\paragraph{Appearance Preservation during Continued Generation}
\label{sec:eval-long-video}

To examine appearance preservation during continued generation, Figure~\ref{fig:lip-color-consistency} compares the generated lip regions with the initial image.
In the illustrated examples, 1.5 shows over-saturated pink or red lip tones, whereas 1.6 remains closer to the initial lip color across the selected frames despite changes in mouth shape.


As shown in Figure~\ref{fig:version-improvements}, the rate of appearance-related issues involving changes to core character attributes decreases from 1.4\% to 0.4\%, while the rate of color or resolution alterations decreases from 3.1\% to 0.0\%. 
Together with the qualitative lip-color examples, these results demonstrate improved preservation of character appearance. 
The overall consistency issue rate also decreases from 19.9\% to 11.6\%, with the rate of speaker-count mismatches declining from 18.5\% to 11.2\%.



\par
\endgroup

\section{Conclusion}
LongCat-Video-Avatar 1.6 demonstrates that targeted post-training improvements can substantially enhance motion naturalness and long-form stability in audio-driven avatar generation. The model addresses key articulation and appearance artifacts, supports video generation exceeding 10 minutes, and achieves favorable results against its predecessor and competing commercial models in Arena evaluations. These results highlight the effectiveness of refining post-training for both perceptual quality and sustained generation.

\newpage 

\section{Contributors and Acknowledgments}

All people are cataloged alphabetically by last name. (\dag) indicates the project leader and (\ddag) indicates the sponsors.

\paragraph{Contributors}\mbox{}\\[1\baselineskip]
\begin{tabular}{p{0.19\textwidth}p{0.19\textwidth}p{0.19\textwidth}p{0.19\textwidth}p{0.19\textwidth}}
Xunliang Cai\ddag & Yiqi Chen & Feng Gao & Xiaoyu Kong & Le Li \\
Hongyu Liu & Xiaoming Wei\ddag & Tianyu Yang & Yong Zhang\dag & Sijie Zhao
\end{tabular}

\paragraph{Acknowledgments}\mbox{}\\[1\baselineskip]
\begin{tabular}{p{0.19\textwidth}p{0.19\textwidth}p{0.19\textwidth}p{0.19\textwidth}p{0.19\textwidth}}
Fengjiao Chen & Yaqin Huang & Siyu Ren & Yingying Wu & \\
\end{tabular}

\bibliographystyle{unsrtnat}
\bibliography{references}  

\clearpage
\appendix

\end{document}